\documentclass[11pt,a4paper]{article}
\usepackage[hyperref]{ranlp2023}
\usepackage{times}
\usepackage{latexsym}

\usepackage{graphicx}

\usepackage{comment}
\usepackage{hyperref}

\usepackage{microtype}
\aclfinalcopy

\title{Measuring Spurious Correlation in Classification: \\"Clever Hans" in Translationese}

\author{Angana Borah\\
  Saarland University \\
  \texttt{anganaborah9@gmail.com} \\ \\ 
  \textbf{Cristina España-Bonet} \\
  Saarland University, DFKI  \\
  \texttt{cristinae@dfki.de} \\\And 
  Daria Pylypenko \\
  Saarland University  \\
  \texttt{daria.pylypenko@uni-saarland.de} \\ \\
  \textbf{Josef van Genabith} \\
  Saarland University, DFKI  \\
  \texttt{josef.van\_genabith@dfki.de} \\}

\author{Angana Borah$^{1}$,
 Daria Pylypenko$^{1}$,
  Cristina España-Bonet$^{2}$,
  Josef van Genabith$^{1,2}$ \\
  $^{1}$Saarland Informatics Campus, Saarland University, Germany\\
  $^{2}$DFKI GmbH, Germany\\
  \texttt{
      anganaborah9@gmail.com, daria.pylypenko@uni-saarland.de,} \\
  \texttt{      \{cristinae,josef.van\_genabith\}@dfki.de}} 

\date{30 Jul 2023}

\begin{document}
\maketitle
\begin{abstract}
Recent work has shown evidence of "Clever Hans" behavior in high-performance neural translationese classifiers, where BERT-based classifiers capitalize on spurious correlations, in particular topic information, between data and target classification labels, rather than genuine translationese signals. Translationese signals are subtle (especially for professional translation) and compete with many other signals in the data such as genre, style, author, and, in particular, topic. This raises the general question of how much of the performance of a classifier is really due to spurious correlations in the data versus the signals actually targeted for by the classifier, especially for subtle target signals and in challenging (low resource) data settings. We focus on topic-based spurious correlation and approach the question from two directions: (i) where we have no knowledge about spurious topic information and its distribution in the data, (ii) where we have some indication about the nature of spurious topic correlations. For (i) we develop a measure from first principles capturing alignment of unsupervised topics with target classification labels as an indication of spurious topic information in the data. We show that our measure is the same as purity in clustering and  propose a "topic floor" (as in a "noise floor") for classification. For (ii) we investigate masking of known spurious topic carriers in classification. Both (i) and (ii) contribute to quantifying and (ii) to mitigating spurious correlations.  

\end{abstract}

\section{Introduction}

The term \emph{translationese} refers to systematic linguistic differences between originally authored texts and translated texts in the same language \cite{gellerstam1986translationese}. Important aspects of translationese have been identified in the linguistic literature \cite{toury1980search, baker1993text, Teich+2012, volansky2015}, including source language interference, over-adherence to target language, simplification, explicitation, and implicitation. Translationese may manifest itself at lexical, syntactic, semantic, and discourse-related levels of linguistic description. While translationese signals are subtle (especially for professional human translation), corpus-based linguistic methods \cite{baker1993text} and machine learning based classification methods \cite{ volansky2015, rabinovich, rubino-etal-2016-information, pylypenko-etal-2021-comparing} are able to reliably distinguish between original and translated texts in the same language, genre, and style. While basic research focuses on identifying and categorizing aspects of translationese, research has also shown that translationese clearly impacts practical cross-lingual tasks that involve translated data \cite{ singh2019, zhang-toral-2019-effect, clark-etal-2020-tydi, artetxe-etal-2020-translation}. Finally, translationese is sometimes regarded as (one of) the final frontier(s) of high-resource machine translation \cite{freitag-etal-2020-bleu, freitag-etal-2019-ape, ni-etal-2022-original}.

In this paper, we focus on translationese classification (into original $O$ and translated $T$ data) using machine learning based approaches. Early work on translationese classification focused on manually engineered and linguistically inspired sets of features (n-grams, POS, discrete LM-based features etc.), using supervised classification models such as decision-trees or support vector machines (SVMs) \cite{baroni, volansky2015, rubino-etal-2016-information}. 

More recently, research focused on feature-and-representation learning neural network methods for translationese classification \cite{sominsky-wintner-2019-automatic, pylypenko-etal-2021-comparing}. \citet{pylypenko-etal-2021-comparing} show that BERT-based approaches outperform handcrafted feature and SVM-based approaches by a large margin (15-20 accuracy points absolute). \citet{amponsah-kaakyire-etal-2022-explaining} show that this performance difference is due to learned features (rather than the classifiers). 

Using Integrated Gradient (IG) based input attribution methods \cite{amponsah-kaakyire-etal-2022-explaining} also show that BERT \cite{devlin-etal-2019-bert} sometimes exploits topic differences between $O$ and $T$ data  as spurious correlations with the target classification labels (original $O$ and translation $T$) as short cuts, rather than "true" translationese signals: the $T$ part of the (Europarl-based) data, translations from Spanish into German, happens to contain mentions of Spanish locations while German originals $O$ tend to mention German location names. Spurious correlations in the data with target classification labels may cause "Clever Hans" behavior \cite{lapuschkin2019unmasking, hernandez2019gazing}, where the classifier picks up accidental patterns in the data correlated with but otherwise unrelated to the classification target, in the case at hand, topic/content differences rather than proper linguistic indicators of translationese.

To the best of our knowledge, to date, we do not know how to measure spurious correlations between topic signals in the data and target classification labels, such as translationese ($O$ and $T$). At the same time, this is an important question, as an answer would allow us to better understand to what extent we can trust a classifier to pick up on information truly relevant to the target classification labels, and to which extent a classifier is exploiting "Clever Hans", i.e. spurious correlations in the data with the target labels. This is especially pertinent with subtle classification targets and challenging low-resource data settings, as in translationese classification: translationese data sets tend to be small, and translationese signals are subtle while competing with many other signals in the data. 

We approach our research question of "measuring spurious topic correlation in the data with respect to target classification labels" from two opposing ends: (i) where we assume no prior knowledge about topics in the data and (ii) where we have some idea about spurious topic signals in the data. We refer to (i) as \textbf{Chasing Unknown Unknowns} (Section \ref{Chasing Unknown Unknowns} below)\footnote{Readers may relate this to a 2002 hearing with the then US Secretary of Defence Donald Rumsfeld. In one scenario we do not know the topics and their impact, the unknown unknowns; in the other we have some indication about a spurious topic but again do not know its impact, the known unknowns.} and (ii) as \textbf{Chasing Known Unknowns} (Section \ref{Chasing Known Unknowns} below). For (i) we use unsupervised topic modeling (LDA and BERTopic) and we develop a measure from first principles that captures the alignment of (unsupervised) topics with target classification labels. Based on this we propose the concept of a "topic floor" in classification, akin to the concept of a "noise floor" in Electronic Engineering. We show that our alignment-based measure is the same as purity with respect to target classes in clustering. Given data, target classification labels and unsupervised topic models, our measure and noise floor provide an upper bound on how much spurious topic information may account for target classification labels. For (ii) we use masking of already identified spurious topic information, such as location names, in the data and measure classification accuracy with masked and unmasked versions of the data, to quantify the impact of the identified source of spurious correlation.

\noindent
Our main contributions include the following:
\vspace{-0.8em}
\begin{enumerate}
    \item We present a measure that, given a data set and target classification labels, quantifies the possible impact of \emph{unknown spurious topic information} on classification. The measure is based on aligning unsupervised topics with the target labels. Based on this we propose the concept of a "topic floor" (akin to "noise floor" in Electronic Engineering) in classification.
    \item We use masking to both quantify and mitigate \emph{known spurious topic information}. 
    \item We present empirical results for topic floor and masking to quantify "Clever Hans" in the translationese data of  \citet{amponsah-kaakyire-etal-2022-explaining}. We use IG attribution to show that in masked settings where known spurious correlations are mitigated, BERT learns features closer to proper translationese.  
\end{enumerate}

\section{Related Work}

\citet{puurtinen2003genre, ilisei2010identification, volansky2015, rabinovich, rubino-etal-2016-information, pylypenko-etal-2021-comparing} train classifiers to distinguish between originally authored and translated data. Many of them explore hand-crafted and linguistically inspired feature sets, manual feature engineering, and a variety of classifiers including Decision Trees and Support Vector Machines (SVMs) and use feature ranking or attribution methods to reason back to particular dimensions of translationese and their importance in the data and classification results. Feature engineering based translationese classification used SVM feature weights \cite{avner:2016, pylypenko-etal-2021-comparing}, decision trees or random forests \cite{ilisei2010identification, rubino-etal-2016-information}, training separate classifiers for each individual feature (or feature sets)
and comparing accuracies \cite{volansky2015, avner:2016}, to explain results. 

More recent research uses feature and representation learning approaches (sometimes augmented with hand-crafted features) based on neural networks \cite{sominsky-wintner-2019-automatic, pylypenko-etal-2021-comparing}. \citet{pylypenko-etal-2021-comparing} shows that feature learning based approaches (e.g. a pretrained BERT based classifier) outperform hand-crafted and feature engineering based approaches (SVM) by as much as 15 to 20 percentage points absolute in classification accuracy. \citet{amponsah-kaakyire-etal-2022-explaining} show that the difference in classification accuracy is due to feature learning rather than the classifiers, and, using Integrated Gradients (IGs) \cite{igsundararajan}, provide evidence that the feature learning methods exploit some spurious correlations with the classification labels in the data, that are clearly not translationese, but topic related cues: the data are  German originals $O$ and translations $T$ (into German) from Spanish, and Spanish place names are highly IG-ranked, given the trained classifiers.


\citet{dutta-chowdhury-etal-2022-towards} use divergence from isomorphism based graph distance measures to show that translationese is visible even in $O$ and $T$ word embedding spaces. While POS features have been used in feature-engineering-based translationese classification, some experiments in \cite{dutta-chowdhury-etal-2022-towards} use POS (instead of the surface words) to mitigate possible topic influences on the graph divergence results. This is an approach we develop further (e.g. in terms of partial masking using NEs) in our work below.  



\section{Data}

Our experiments use the monolingual German dataset from the Multilingual Parallel Direct Europarl (MPDE) corpus \cite{amponsah-kaakyire-etal-2021-rely} consisting of 42k paragraphs, with half of the paragraphs German (DE) originals (below we call this $O$) and the other half translations (below we call this $T$) from Spanish (ES) to German. The average length (in terms of tokens) per training example is 80. Like all of MPDE, the DE-ES subset contains only data from before 2004, since for post-2004 data, it may not be known whether or not the source language SL is already the result of a translation \cite{bogaert2011absolute}. While this limits the amount of data, it ensures that the $O$ and $T$ data are clearly identifiable and "pure" $O$ or $T$. Both $O$ and $T$ are German, but $T$ is German translated from Spanish, and both coming from MPDE ensure that they are the same Europarl genre and style.


\section{Chasing Unknown Unknowns}\label{Chasing Unknown Unknowns}

In this section, we assume that we have no prior information about topics and their distribution in the data. Because of this, we use unsupervised topic modeling. We develop a measure that checks whether, and if so to what extent, the topics established in this fashion \emph{align} with the target classes $O$ and $T$ in our data. The measure quantifies to which extent topic is a giveaway for translationese. 

\subsection{How to Measure Topic Bias Relevant to Translationese Classification?}

The goal is to investigate the amount and distribution of topic signal in $O$ and $T$ data, that could be used as a spurious signal in translationese (i.e. $O$ and $T$) classification. As initially we do not know anything about possible topics and their distribution in the data, we use standard approaches to unsupervised topic modeling, like  Latent Dirichlet Allocation (LDA) \cite{lda} and BERTopic \cite{grootendorst2022bertopic}. Both LDA and BERTopic will cluster our data into classes, i.e. topics. How can we measure whether the topics established by the topic model are potentially relevant to translationese classification? Topics are relevant to $O$ and $T$ translationese classification if the paragraphs in each of the topics are either mostly $O$ paragraphs or if they are mostly $T$ paragraphs, in other words if topics are well \emph{aligned} to either $O$ or $T$. If this is the case, a translationese classifier may learn to use topic, rather than proper translationese signals (or a mix of both) in translationese classification. To give a simple (and extreme) example, suppose we take the union\footnote{As $O$ and $T$ paragraphs are disjoint, this is the same as their concatenation.} of $O$ and $T$ and cluster the union using LDA into, say, two topics (classes) \emph{top}$_1$ and \emph{top}$_2$. If (and this is the extreme case) \emph{top}$_1$ = $O$ and \emph{top}$_2$ = $T$ (or vice versa, i.e. \emph{top}$_1$ = $T$ and \emph{top}$_2$ = $O$), then topic perfectly predicts $O$ and $T$. We would like our measure to capture this, and we would like the measure to be symmetric, i.e. give the same result no  matter whether \emph{top}$_1$ = $O$ and \emph{top}$_2$ = $T$ or vice-versa. Now consider another (extreme) case: lets say \emph{top}$_1$ is half $O$ and half $T$, with \emph{top}$_2$ the same but with the other halfs of $O$ and $T$. In this case topic is not able to distinguish between $O$ and $T$ (beyond chance). What about cases in between the two extreme cases? Lets say, \emph{top}$_1$ is 3/4 $O$ and 1/4 $T$, and therefore \emph{top}$_2$ is 1/4 $O$ and 3/4 $T$ (or vice versa - swap \emph{top}$_1$ and \emph{top}$_2$). In this case topics \emph{top}$_1$ and \emph{top}$_2$ are pretty good indicators of $O$ and $T$, and a translationese classifier may pick up on topic signals rather than just translationese proper.   


To design a measure that captures the relevance of topic classes to (binary) translationese classification, we need our measure to be symmetric, generalize to more than 2 topic classes, and factor in possible $O$ and $T$ class imbalance. To keep things simple\footnote{We could design an entropy-based measure of the distribution of topic classes with respect to $O$ and $T$, factor in classification probabilities of LDA etc.}, here we present a straightforward and easy to use measure we call \emph{alignment of topic top}$_i$ \emph{with}  $O$ \emph{and} $T$, denoted 

\begin{equation}
   \textrm{\emph{align}}_{O,T}(\textrm{\emph{top}}_i) 
\end{equation}

\noindent with the majority class $O$ or $T$ covered by \emph{top}$_i$ (whatever it is for a given \emph{top}$_i$) given the benefit of the doubt as the "correct" translationese class. We assume that \emph{Data} = $O \cup T$, $O \cap T = \emptyset$, $\bigcup_{i=1}^{n} \textrm{\emph{top}}_i$ = \emph{Data} and  $\bigcup_{i \neq j} \textrm{\emph{top}}_i \cap \textrm{\emph{top}}_j = \emptyset$, i.e. topic partitions our data, as does $O$ and $T$. With this

\begin{equation}\label{1}
    \textrm{\emph{align}}_{O,T} (\textrm{\emph{top}}_i) = \frac{ \max(|\textrm{\emph{top}}_i \cap O|, |\textrm{\emph{top}}_i \cap T|)}{|\textrm{\emph{top}}_i|}
\end{equation}



\noindent $\max(\cdot,\cdot)$ makes the measure symmetric. Given \emph{align}$_{O,T} (\textrm{\emph{top}}_i)$, the weighted average is simply:

\begin{equation}
    \textrm{\emph{avg\_align}}_{O,T}(\textrm{\emph{tops}}) = \sum_{i=1}^{n} w_i \times \textrm{\emph{align}}_{O,T}(\textrm{\emph{top}}_i) 
\end{equation}

\noindent where a weight $w_i = |\textrm{\emph{top}}_i|/|\textrm{\emph{Data}}|$ 
is just the proportion of paragraphs in topic \emph{top}$_i$ divided by the total number of paragraphs in the data. 


It is easy to see that the definition generalizes to $n$ topic classes \emph{top}$_1$ to \emph{top}$_n$, that it adapts to different \emph{top}$_i$ topic sizes as well as the class imbalance between $O$ and $T$\footnote{To see this, note that as $O$, $T$ partition the data and as $\bigcup\textrm{\emph{top}}_i$ also partition the data, if, let us say, $O \ll T$ and for some \emph{top}$_i$ = $O$, then there is nothing of $O$ left to any of the other \emph{top}$_{j \neq i}$.  Note that in our data we have $|O|\approx |T|$).}. \emph{align}$_{O,T}$(\emph{top}$_i$) $\in$ [0.5, 1] where  \emph{align}$_{O,T}$(\emph{top}$_i$) = 1 signals perfect alignment of topic \emph{top}$_i$ with one of $O$ or $T$, and that \emph{align}$_{O,T}$(\emph{top}$_i$) = 0.5 signals that \emph{top}$_i$ is maximally undecided with respect to $O$ and $T$. And the same for \emph{avg\_align}$_{O,T}$(\emph{top}).

Our alignment-based measure\footnote{One of our reviewers suggested we compare our measure to existing cluster quality measures.} is in fact the same as cluster \emph{purity} \cite{clustering} defined as 

\begin{equation}
   \frac{1}{M} \sum_{clu \in Cluster} \max_{cla \in Class} (clu\cap cla) 
\end{equation}

\noindent where $M$ is the size of the data, $Cluster$ and $Class$ the set of clusters and classes, respectively. With this we have


\[ \textrm{\emph{avg\_align}}_{Class}(\textrm{\emph{Cluster}}) \]
 \[= \sum_{clu \in Cluster} w_{clu} \times  \textrm{\emph{align}}_{Class}(clu)\]

\[ = \sum_{clu \in Cluster} \frac{|clu|}{M} \times \frac{\max_{cla \in Class}(clu \cap cla)}{|clu|} \]
 \begin{equation}
= \frac{1}{M} \sum_{clu \in Cluster} \max_{cla \in Class} (clu\cap cla) 
\end{equation}


\subsection{Experiments}

We use LDA \cite{lda} as our main unsupervised topic model, as it provides a standard and well-understood baseline\footnote{We use the \emph{Gensim} \cite{gensim} implementation of \emph{Mallet} LDA \cite{mallet}.}. LDA makes two key assumptions: (1) documents are a mixture of topics, and (2) topics are a mixture of words. LDA generates a document-term matrix (DTM), where each document is represented by a row and the terms (words) corresponding to each document are represented by the columns. The DTM is decomposed into a document-topic matrix and a topic-word matrix. LDA assigns every word to a latent topic (\emph{top}$_i$) through iteration, computing a topic word distribution ($\theta$) in the data. To build this distribution, LDA uses two parameters: $\alpha$ which controls the per-document topic distribution, and $\beta$ (the Dirichlet parameter) which controls the per-topic word distribution.  LDA requires us to specify the number of topics $n$ in advance. In our experiments we explore $n$ over three orders of magnitude, roughly doubling $n$ at each step, starting with $n=2$ and going up to $n=500$.

As LDA requires us to specify $n$ in advance, we also use BERTopic \cite{grootendorst2022bertopic}, which can find an optimal $n$ given the data\footnote{We partly do this as a sanity check to assess whether our steps increasing $n$ across three orders of magnitude for LDA missed an important region of number of topics.}. By default, BERTopic utilizes contextual sentence embeddings (SBERT), dimensionality reduction (UMAP), clustering (HDBSCAN), tokenizing (CountVectorizer), and a weighing scheme (c-TFIDF) to perform topic modeling. We choose the embedding model 'T-Systems-onsite/cross-en-de-roberta-sentence-transformer' from \emph{Huggingface} \cite{wolf-etal-2020-transformers} and the defaults for all other modules.  

For LDA, we explore a number of topics $n$ with $n=2, 5, 10, 20, 30, 50, 100, 200, 300, 400$ and $500$, over three orders of magnitude. BERTopic returns 207 topics\footnote{BERTopic is stochastic due to UMAP and returns a different number of topics for each run, however, the differences are small.}. 

For each document (here paragraph), we use the highest probability LDA or BERTopic topic assigned to the document to label the document. A topic is then represented by the set of documents labeled with the topic. For each topic \emph{top}$_i$, we compute how well the topic is aligned with $O$ and $T$, i.e. we compute \emph{align}$_{O,T}$(\emph{top}$_i$), and the weighted average over the topics: \emph{avg\_align}$_{O,T}$(\emph{top}$_1$, ..., \emph{top}$_n$). 

\subsection{Results}

\begin{figure}
\centering
  \includegraphics[width=7.6cm]{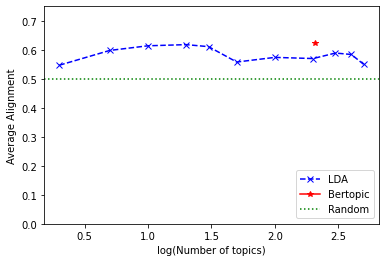}
\caption{Average Topic - Target Classification Alignment \emph{avg\_align}$_{O,T}$(\emph{top}$_1$, ..., \emph{top}$_n$) for LDA and Bertopic}
\label{fig:weighted}

\end{figure}

We plot \emph{avg\_align}$_{O,T}$(\emph{top}$_1$, ..., \emph{top}$_n$) in Fig. \ref{fig:weighted}, varying $n$ from $2$ to $500$, at each step roughly doubling $n$ for LDA, and with $n=207$ for and as determined by BERTopic. An average alignment of 0.5 (the dashed green line) shows topics maximally undecided with respect to $O$ and $T$, while a score of 1 indicates perfect alignment where topics completely predict $O$ and $T$. Fig. \ref{fig:weighted} shows topic alignment with $O$ and $T$ in the range of 0.55 to 0.62, depending on $n$, with BERTopic achieving the overall highest score of 0.62 at $n=207$. For LDA, scores are highest (0.611 - 0.618) for $n = 10, 20$, and $30$. For good choices of topic numbers $n$, both LDA and BERTopic topics are able to predict $O$ and $T$ (i.e. translationese) by close to 0.62. 

\subsection{Discussion and Interpretation: the "Topic Floor" in Classification}

This is an interesting and perhaps somewhat surprising result, but what exactly does it mean? There are two important caveats: 

First, the fact that topic is able to predict $O$ and $T$ by close to 0.62 in the data does not necessarily mean (i.e. prove) that a high-performance BERT translationese classifier, such as the one presented in \cite{pylypenko-etal-2021-comparing}, necessarily uses spurious topic information aligned with $O$ and $T$ in the data. At the same time, however, it cannot be ruled out. As a sanity check we tested how well a BERT classifier can learn to predict LDA topic classes for $n = 2, 10, 20$ and $30$ and for BERTopic's $207$. The results are $0.83, 0.64, 0.42, 0.44$ and $0.57$ (all acc. and well above the largest class baseline). Given how well BERT-based classifiers can pick up patterns in the data, it is prudent to assume that BERT will be sensitive to and use topic signals spuriously aligned with $O$ and $T$. 

Second, we cannot at this stage completely rule out (other than perhaps through laborious manual inspection) that some LDA or BERTopic topics may in fact reflect genuine rather than spurious signals. LDA, e.g., uses lexical information, and perhaps some such information is a genuine translationese signal (unlike the place names clearly identified as a spurious topic signal in \cite{amponsah-kaakyire-etal-2022-explaining}), such as e.g. certain forms of verbs (see Section \ref{Chasing Known Unknowns} on the Known Unknowns).

The two caveats are aspects of the "unknown unknowns" we are chasing in this part of the paper. We have a clear indication that topic aligns with and hence can predict $O$ and $T$ in our data up to 0.62. There are very good reasons (but no proof) to assume that it is likely that a high-performance BERT classifier may use this, while we cannot completely rule out that some of the supposedly spurious topic signal may actually be genuine translationese. Given this, 0.62 is an \emph{upper bound} of how well topic may predict $O$ and $T$ in our data. We may be well advised to take inspiration from the concept of a "noise floor" in Electronic Engineering. The noise floor is the hum and hiss (of a circuit) due to the components when there is no signal, and below which we cannot identify a signal. Given our findings, perhaps we should regard the 0.62 topic alignment with $O$ and $T$ in our data as a "topic floor" for translationese classification. This is in fact the recommendation we take from our work: instead of using 0.5 as a random baseline for our (roughly) balanced binary translationese data set, we should require 0.62 as established by the topic alignment experiments as a safe(r) baseline. Put differently, given our data we cannot really be sure about $O$ and $T$ classification results $\leq 0.62$. Suffice it to say, even with a 0.62 baseline, (most of) the classifiers presented in \cite{pylypenko-etal-2021-comparing, amponsah-kaakyire-etal-2022-explaining} easily surpass that baseline (with acc $\geq 0.9$).



\section{Chasing Known Unknowns}\label{Chasing Known Unknowns}

In this section, we assume that we have some knowledge about spurious topics in our data and that we want to both quantify and mitigate this spurious topic information in translationese classification. The difference in classification accuracy between a classifier that has access and one that does not have access to spurious topic information quantifies the impact of the spurious topic information in question. Using IG, \citet{amponsah-kaakyire-etal-2022-explaining} show that high-performance BERT-based classifiers use location names (Spanish names in the $T$, and German names in the $O$ data) in the classification, clearly not a proper translationese but a spurious topic signal. Similarly, named entities (NEs) are often highly ranked in LDA topics. Therefore, the most straightforward approach towards mitigating specific spurious topic information in translationese classification is identifying NEs and masking them in the data. 


\subsection{Named Entity Recognition on Europarl}

We focus on a scenario where we identify (and later mask) NEs automatically, rather than manually. While automatic NER is noisy, unlike potentially high-quality manual NER, it scales and constitutes a realistic application scenario. To assess NER performance, we experiment with a number of SOTA NER models, namely SpaCy \cite{spacy2}, FLERT \cite{flert}, multilingual-BERT \cite{bert}, XLM-RoBERTa \cite{xlm}, DistilBERT \cite{distilbert}, and BERT-German \cite{gbert}, comparing F1, precision and recall for each of the models against a gold standard NE tagged dataset \cite{agerri-etal-2018-building}.  The gold standard consists of 800 sentences from the Europarl German data manually annotated following the ConLL 2002 \cite{conll2002} and 2003 \cite{conll2003} guidelines, with a total of 433 named entities.

\begin{table}
\small
\centering
\begin{tabular}{|p{2.3cm}|p{1.2cm}|p{1cm}|p{1.2cm}|}
\hline
\textbf{NER model} & \textbf{Precision} & \textbf{Recall} & \textbf{F1-score}\\
\hline
SpaCy & 0.26 & 0.56 & 0.35 \\
FLERT &  0.39 & 0.35 & 0.37 \\
mBERT-large &  0.33 & 0.31 & 0.32 \\
XLM-R-base & 0.20 & 0.33 & 0.25 \\
XLM-R-large & 0.19 & 0.31 & 0.24 \\
DistilBERT-large & 0.34 & 0.31 & 0.32 \\
BERT-German & 0.65 & 0.42 & 0.52 \\ 
\hline
\end{tabular}
\caption{\label{ner_comparison}
Comparison table for NER models on the gold standard dataset from \cite{agerri-etal-2018-building}
}
\end{table}

Table ~\ref{ner_comparison} shows that BERT-German (a fine-tuned version of bert-base-multilingual-cased on the German WikiANN dataset) has the highest precision, second-highest recall and the highest F1 (0.52) among all the NER models. Hence, we choose BERT-German for all our NER experiments.



\subsection{Translationese Classification}
To quantify the impact of NE-based spurious topic information on translationese classification, we modify our data by masking NEs (Section \ref{nemdesc}), and, in a separate experiment, we explore full masking of the data using POS (Section \ref{POS Full Masking}). 

Following \cite{pylypenko-etal-2021-comparing}, we use multilingual BERT \cite{bert} (BERT-base-multilingual-uncased) and fine-tune BERT on the training set of our data set using the \emph{Huggingface} library. We use a batch size of 16, a learning rate of $4 \cdot 10^{-5}$, and the Adam optimizer with $\epsilon = 1 \cdot 10^{-8}$. 

We compare our models with the BERT model reproduced from \cite{pylypenko-etal-2021-comparing}: a pre-trained BERT-base model (12 layers, 768 hidden dimensions, 12 attention heads) fine-tuned on translationese classification using unmasked data. 



\subsubsection{Named Entity Masking} \label{nemdesc}

We use Bert-German to replace NEs with one of three course-grained NE-type tags: [LOC], [PER], and [ORG]. For example, the unmasked string "\emph{John will go to Berlin.}" is NE masked as "[PER] \emph{will go to} [LOC]". In our train-dev-test sets, we have 202036, 42072, and 43489 NEs. We carry out experiments with four train-test configurations: masked-masked, unmasked-masked, masked-unmasked, and the original unmasked-unmasked. For each of these configurations, we fine-tune BERT to the specifics of the training set (i.e. masked or unmasked). We supply the three NE-type tags as special additional tokens to the BERT tokenizer to ensure that they are consistently represented by their NE-type token (and that no sub-word splitting is applied to NE-type tokens). If NEs are  responsible for spurious topic information in translationese classification, we expect masking NEs to mitigate spurious correlations in the data and to result in reduced translationese classification accuracies, allowing us to quantify this aspect of spurious topic correlations.

\subsubsection{Part-Of-Speech (POS) Full Masking}\label{POS Full Masking}


To analyze BERT's performance on fully delexicalized data, we use the finer POS tagger from SpaCy \cite{spacy2} which utilizes the TIGER Treebank \cite{brants2004tiger}\footnote{\url{https://github.com/explosion/spaCy/blob/master/spacy/glossary.py} (last accessed 11 Aug, 2023)}. 
Given: "\emph{Jetzt solle erneut ein Antrag gestellt werden .}", the POS tag sequence is: \textsc{"ADV VMFIN ADJD ART NN VVPP VAINF \$."}.
We pre-train BERT on POS-tagged data from 3\% of the German Wikipedia dump (1.5 million sentences) on the BertforMaskedLM objective for 2 epochs. We use BertWordPieceTokenizer for tokenization. To adjust the BERT model to the small vocabulary, we use only 6 encoder layers (instead of 12), a learning rate of $5.10^{-6}$, and the Adam optimizer with  $\epsilon = 1.10^{-8}$. 
We use the POS-pre-trained model and fine-tune it with the POS-tagged monolingual German dataset from the MPDE corpus \cite{amponsah-kaakyire-etal-2021-rely} (same fine-tuning parameters as other experiments).  

\subsection{Integrated Gradients (IG)}
\citet{amponsah-kaakyire-etal-2022-explaining} use IG attribution scores to show that BERT utilizes spurious correlations in the data, for example, German $T$ data translated from Spanish contain mentions of Spanish geographical areas, such as  'Spanien', 'Barcelona' etc. as top tokens identified by IG.
Here we use IG on BERT trained on masked data, and compute the top tokens with the highest attribution scores on average across the masked test sets. 
We also compute the top POS tags by performing IG on BERT trained on fully POS-tagged data.

\subsection{Results}

\begin{table}[ht]
\small
\centering
\begin{tabular}{|p{1.7cm}|p{3cm}|p{1.7cm}|}
\hline
\textbf{Train-Test} & \textbf{Test Set Acc (\%)} & \textbf{95\% CI} \\
\hline
 m-m & 0.89$\pm$0.00 & [0.88,0.89] \\
 m-u & 0.89$\pm$0.00 &	[0.89,0.90] \\
 u-m & 0.90$\pm$0.00 &	[0.90,0.91]\\
 u-u & 0.92$\pm$0.00 & [0.91,0.93]\\ \hline
\end{tabular}
\caption{\label{baseline_and_new_new}
NE masked experiments pretrained-BERT-ft Acc(uracy); CI(Conf. Interval); m(asked), u(nmasked).
}
\end{table}

\begin{table}[ht]
\small
\centering
\begin{tabular}{|p{4cm}|p{1.8cm}|}
\hline
\textbf{Test Set Acc (\%)} & \textbf{95\% CI} \\
\hline
0.78 $\pm$ 0.00 & [0.77, 0.79] \\ \hline
\end{tabular}
\caption{\label{pos}
POS-masked experiments POS BERT fine-tuned with TIGER Treebank tags, Acc(uracy); CI(Conf. Interval)
}
\end{table}

\begin{table}[ht]
\centering
\small
\begin{tabular}{|l|l|r|l|r|}
\hline
{} & \multicolumn{2}{l|}{\textbf{Translationese}} & \multicolumn{2}{l|}{\textbf{Original}} \\
\hline
 &             Token &   AAS &     Token &   AAS \\
\hline
1    &          besuchte &  0.61 &           • &  0.83 \\
2    &         entdeckte &  0.60 &       alpen &  0.69 \\
3    &   veroffentlichte &  0.53 &         apo &  0.66 \\
4    &          gehorten &  0.51 &     profits &  0.63 \\
5    &            fuhrte &  0.47 &      \#\#nova &  0.59 \\
6    &           nominal &  0.46 &       super &  0.49 \\
7    &           benutzt &  0.46 &       \#\#bud &  0.48 \\
8    &              tari &  0.45 &      \#\#ndus &  0.46 \\
9    &             starb &  0.44 &    \#\#enland &  0.46 \\
10   &              eman &  0.43 &     \#\#hutte &  0.45 \\
11   &             loste &  0.39 &    digitale &  0.45 \\
12   &          planeten &  0.39 &         ros &  0.45 \\
13   &           geboren &  0.38 &  population &  0.43 \\
14   &  veroffentlichten &  0.38 &         pla &  0.43 \\
15   &             neige &  0.37 &     express &  0.42 \\
16   &           schrieb &  0.37 &     \#\#vagen &  0.40 \\
17   &          priester &  0.36 &       stahl &  0.40 \\
18   &        scheiterte &  0.36 &          ez &  0.40 \\
19   &             genus &  0.35 &      stands &  0.40 \\
20   &       territorium &  0.35 &       \#\#nog &  0.39 \\
\hline
\end{tabular}
\caption{\label{tab:ig1}Top-20 tokens with highest IG average attribution score (AAS) for the NE-masked test set.}
\end{table}

\begin{table}[ht]
\small
\centering
\resizebox{\columnwidth}{!}{%
\begin{tabular}{|l|l|r|l|r|}
\hline
{} & \multicolumn{2}{l|}{\textbf{Translationese}} & \multicolumn{2}{l|}{\textbf{Original}} \\
\hline
&             Token &   AAS &     Token &   AAS \\
\hline
1    &           APPO (ADP) &  0.32 &      ADV (ADV) &  0.21 \\
2    &          PRELS (PRON) &  0.19 &       . (PUNCT) &  0.12 \\
3    &           KOUI (SCONJ) &  0.15 &    TRUNC (X) &  0.06 \\
4    &         PPOSAT (DET) &  0.14 &     ADJD (ADJ) &  0.05 \\
5    &         PRELAT (DET) &  0.12 &       FM (X) &  0.04 \\
6    &            PIS (PRON) &  0.12 &    PROAV (ADV) &  0.03 \\
7    &           PPER (PRON) &  0.11 &      PDS (PRON) &  0.03 \\
8    &           PDAT (DET) &  0.11 &    VVIZU (VERB) &  0.02 \\
9    &          VVFIN (VERB) &  0.11 &   PTKANT (PART) &  0.02 \\
10   &          VMFIN (VERB) &  0.10 &    PTKZU (PART) &  0.01 \\
\hline
\end{tabular}
}
\caption{\label{tab:ig2}Top-10 tokens with highest IG average attribution score (AAS) for the POS-tagged test set (TIGER Treebank tags). Corresponding UPOS tags are given in braces. We use the conversion table from \url{https://universaldependencies.org/tagset-conversion/de-stts-uposf.html} (last accessed 11 Aug, 2023).}
\end{table}

\subsubsection{Results NE Masking}

Table~\ref{baseline_and_new_new} shows test set accuracies for the NE masking experiments outlined in Section \ref{nemdesc}. We use Bootstrap Resampling, with 100 samples and 95\% confidence intervals. Results (u-u against all others) are statistically significant. Consistent with expectation, under all training-test data conditions, masking NE-related information lowers classification results. Compared to the \cite{amponsah-kaakyire-etal-2021-rely} unmasked-unmasked baseline, the performance drop is between 0.026-0.032 points absolute. In absolute terms, the performance drop incurred in mitigating spurious topic information in terms of NEs masking is visible but small, in the order of 3 to 4 \% points absolute, if classification accuracy is expressed as \% points. This indicates that this type of spurious topic information and the ensuing "Clever Hans" is a small part of the strong BERT translationese classification performance. 

\subsubsection{Results Full POS Masking}

Table~\ref{pos} shows that translationese classification results on fully de-lexicalized POS-masked data are much lower than for NE-masked data\footnote{For fully PoS-masked data it does not make sense to report mixed train-test conditions.}. In this regime, BERT is missing much valuable information. At the same time, the classification accuracy of almost 0.78 shows that BERT is able to pick up on morpho-syntactic aspects of translationese. We also performed an analogous experiment with Universal POS tags (see Appendix, section \ref{Full_UPOS_masking}), and obtained accuracy of almost 0.77.

\subsubsection{Integrated Gradients NE and POS}

Table~\ref{tab:ig1} shows the top-20 IG token attributions for $O$ and $T$ data in the masked-masked condition. Unlike \cite{amponsah-kaakyire-etal-2022-explaining} the "translationese" column does not show any Spanish or other place names. At the same time the "original" column still contains a few location tokens (or possible subwords of location tokens), such as "alpen", "\#\#enland", "ez" (as in \cite{amponsah-kaakyire-etal-2022-explaining}), confirming the fact that automatic NER is not perfect (see P, R and F1 scores in Table~\ref{ner_comparison}).

It is interesting to note that hand-in-hand with the reduction of spurious NE-based cues, many of the observations from \cite{amponsah-kaakyire-etal-2022-explaining} are confirmed and in fact come to the fore. In the $T$ class we observe an even stronger presence of verbs in Präteritum form (this time not only regular, but also irregular verbs), which \citet{amponsah-kaakyire-etal-2022-explaining} link to the fact that translators might have preferred to use a more written style while translating the transcribed speeches.

Table~\ref{tab:ig2} presents the top-10 IG attributed tokens for the POS-tagged test set, for BERT trained on POS-tagged data.
Interestingly, the top tags are APPO (postpositions) for the translationese class $T$, and ADV (adverbs) for the originals class $O$, which confirms findings in \citep{pylypenko-etal-2021-comparing} who show that relative frequencies of adverbs and adpositions are among the highest-ranked features that correlate with predictions of various translationese classification architectures, including BERT (even though their experiment is performed in the multilingual setting, and not on just German data like ours). They also show that the ratio of determiners is an important feature, and we see many tags corresponding to this category in our list: PPOSAT (attributive possessive pronouns), PRELAT (attributive relative pronouns), PDAT (attributive demonstrative pronouns), etc. The results for UPOS tags are similar (see Appendix, section \ref{Full_UPOS_masking}).





\section{Conclusion}

We present a measure that, given a data set and target classification labels, quantifies the possible impact of \emph{unknown spurious topic information on classification}. The measure is based on aligning unsupervised topics with target labels and is equivalent to \emph{purity} in clustering. We propose the concept of a "topic floor" (akin to "noise floor") as an upper bound of the impact of spurious topic information on classification in classification. We use masking to quantify and mitigate \emph{known spurious topic information}. We present empirical results for topic floor and masking to quantify "Clever Hans" in the translationese data of  \cite{amponsah-kaakyire-etal-2022-explaining}. We use IG attribution to show that in masked settings where known spurious correlations are mitigated, BERT learns features closer to proper translationese.  

\section*{Acknowledgements}

We are grateful for the insightful and critical feedback from our three reviewers. The research was funded by the Deutsche Forschungsgemeinschaft (DFG)-ProjectID 232722074-SFB 1102.

\bibliographystyle{acl_natbib}
\bibliography{anthology,ranlp2023}

\appendix
\section{Appendices} 
\label{sec:appendix} 

\subsection{Topics by LDA and Bertopic}
Here we take a closer look at LDA and Bertopic topics. While we do not find much evidence for geographic LDA topics, we do find quite a few geographic BERTtopic topics, for example, Topic 10 consists of word tokens "t\"urkei, t\"urkischen, t\"urkische, kriterien, helsinki, da\ss, politischen, kurdischen, menschenrechte, die", Topic 14: "pal\"astinensischen, israel, arafat, israelischen, pal\"astinensische, sharon, autonomiebehörde, pal\"astinenser, frieden, israels", Topic 23: "kuba, kubanischen, kubaner, kubas, kubanische, dissidenten, volk, castro, cotonou, havanna" predominantly consist of geographical terms. 

\subsection{Topic Classification}
To understand if BERT is able to learn the topics identified by the topic modeling experiments, we perform topic classification by finetuning pre-trained-BERT on the topics found by LDA and BERTopic. 


We use a similar ratio for each topic as the train:dev:test (29580:6336:6344) ratio for the translationse classification experiments. 
\begin{table}[ht]
\small
\centering
\begin{tabular}{|p{1.3cm}|p{1.8cm}|p{1.5cm}|p{1.3cm}|}
\hline
\textbf{n} & \textbf{Test Set Acc (\%)} & \textbf{95\% CI} & \textbf{Baseline Acc}\\
\hline
2 & 0.832$\pm$0.00 & [0.83,0.84] & 0.50 \\
10 & 0.636$\pm$0.01 &	[0.62,0.64] &  0.18 \\
20 & 0.417$\pm$0.00 &	[0.41,0.42] & 0.13 \\
30 & 0.442$\pm$0.00 & [0.44,0.45] & 0.11 \\ 
207 (BT) & 0.569$\pm$0.00 & [0.56,0.57] & 0.002 \\ \hline
\end{tabular}
\caption{\label{top_class}
Topic Classification experiments pretrained-BERT-ft Acc(uracy); n(umber of topics), CI(Conf. Interval), BT (BERTopic).
}
\end{table} \newline

Table ~\ref{top_class} shows the topic classification results for the topics output by LDA and BERTopic. We also show baseline accuracies, when the model only predicts the largest class.

\subsection{Full UPOS Masking}
\label{Full_UPOS_masking}

Apart from full masking with detailed tags from the TIGER Treebank, we also explore full masking with the more general Universal POS tags. BERT was pre-trained and fine-tuned on the POS-tagged data in the same way as described in section \ref{POS Full Masking} for the TIGER tags. The translationese classification accuracy (Table \ref{upos}) is slightly lower than for the detailed tags (Table \ref{pos}). 

\begin{table}[ht]
\small
\centering
\begin{tabular}{|p{4cm}|p{1.8cm}|}
\hline
\textbf{Test Set Acc (\%)} & \textbf{95\% CI} \\
\hline
0.768 $\pm$ 0.00 & [0.76, 0.77] \\ \hline
\end{tabular}
\caption{\label{upos}
POS-masked experiments POS BERT fine-tuned with UPOS tags, Acc(uracy); CI(Conf. Interval)
}
\end{table}

IG results (Table \ref{tab:ig3}) show patterns similar to those for the detailed tags (Table \ref{tab:ig2}): PRON, DET and SCONJ for translationese; X, ADV, PART and ADJ for originals.

\begin{table}[ht]
\small
\centering
\begin{tabular}{|l|l|r|l|r|}
\hline
{} & \multicolumn{2}{l|}{\textbf{Translationese}} & \multicolumn{2}{l|}{\textbf{Original}} \\
\hline
&             Token &   AAS &     Token &   AAS \\
\hline
1    &           PRON &  0.12 &        X &  0.31 \\
2    &          PUNCT &  0.08 &      ADV &  0.16 \\
3    &          CCONJ &  0.07 &     PART &  0.07 \\
4    &            DET &  0.06 &      AUX &  0.05 \\
5    &          SCONJ &  0.04 &     VERB &  0.03 \\
6    &           NOUN &  0.02 &      NUM &  0.02 \\
7    &            &   &    PROPN &  0.02 \\
8    &           &   &   ADJ &  0.02 \\
9    &          &   &  NOUN &  0.02 \\
\hline
\end{tabular}
\caption{\label{tab:ig3}Top-10 tokens with highest IG average attribution score (AAS) for the POS-tagged test set (UPOS tags).}
\end{table}

\end{document}